\documentclass[a4paper,fleqn]{cas-dc}

\usepackage[numbers]{natbib}
\usepackage{adjustbox}
\usepackage{ctable}
\usepackage{changepage}
\usepackage{colortbl}
\usepackage{color,soul}
\usepackage{amsmath,amsfonts,amssymb}
\usepackage{algorithmic}
\usepackage{algorithm}
\usepackage{array}
\usepackage[caption=false,font=normalsize,labelfont=sf,textfont=sf]{subfig}
\usepackage{textcomp}
\usepackage{stfloats}
\usepackage{url}
\usepackage{verbatim}
\usepackage{graphicx}
\usepackage[utf8x]{inputenc}
\usepackage{xcolor}
 \usepackage{hyperref}
 \hypersetup{
     colorlinks = true,
     linkbordercolor = {},
     allcolors  = blue
 }

\def\tsc#1{\csdef{#1}{\textsc{\lowercase{#1}}\xspace}}
\tsc{WGM}
\tsc{QE}
\tsc{EP}
\tsc{PMS}
\tsc{BEC}
\tsc{DE}
\begin{document}
\let\WriteBookmarks\relax
\def\floatpagepagefraction{1}
\def\textpagefraction{.001}

\shorttitle{}

\shortauthors{Faruqui et al.}

\title [mode = title]{A Model Predictive Control Functional Continuous Time Bayesian Network for Self-Management of Multiple Chronic Conditions}                      
\author{Syed Hasib Akhter Faruqui\textsuperscript{a},
Adel Alaeddini\textsuperscript{b},
Jing Wang\textsuperscript{c},
Susan P Fisher-Hoch\textsuperscript{d},
Joseph B Mccormick\textsuperscript{d},
and Julian Carvajal Rico\textsuperscript{b}
}
\address[1]{Department of Radiology, Northwestern University, Chicago, IL 60611}
\address[2]{Department of Mechanical Engineering, The University of Texas at San Antonio, San Antonio, TX 78249}
\address[3]{College of Nursing, Florida State University, Tallahassee, FL 32306}
\address[4]{School of Public Health Brownsville, The University of Texas Health Science Center at Houston, TX 78520}

\nonumnote{Corresponding Author: Adel Alaeddini (e-mail: \url{adel.alaeddini@utsa.edu}).}

\begin{abstract}
Multiple chronic conditions (MCC) are one of the biggest challenges of modern times. The evolution of MCC follows a complex stochastic process that is influenced by a variety of risk factors, ranging from pre-existing conditions to 
modifiable lifestyle behavioral factors (e.g. diet, exercise habits, tobacco use, alcohol use, etc.) to non-modifiable socio-demographic factors (e.g., age, gender, education, marital status, etc.).
%%%%%%%%%%%%%%%%%%%%%%%
People with MCC are at an increased risk of new chronic conditions and mortality. 
%%%%%%%%%%%%%%%%%%%%%%%%
% Several studies have identified the protective effect of a healthy lifestyle on chronic diseases. Furthermore, a healthy lifestyle imparts a life expectancy free of complications of MCC. 
% Thus, the need for proactive measures to better manage and control the emergence and progression of MCC via lifestyle behaviors. 
%%%%%%%%%%%%%%%%%%%%%%%%%%%%%
This paper proposes a model predictive control functional continuous time Bayesian network, an online recursive method to examine the impact of various lifestyle behavioral changes on the emergence trajectories of MCC, and generate strategies to minimize the risk of progression of chronic conditions in individual patients. 
%%%%%%%%%%%%%%%%%%%%%%%%%%%%%
The proposed method is validated based on the Cameron county Hispanic cohort (CCHC) dataset, which has a total of 385 patients. The dataset examines the emergence of 5 chronic conditions (diabetes, obesity, cognitive impairment, hyperlipidemia, and hypertension) based on four modifiable risk factors representing lifestyle behaviours (diet, exercise habits, tobacco use, alcohol use) and four non-modifiable risk factors, including socio-demographic information (age, gender, education, marital status). 
%%%%%%%%%%%%%%%%%%%%%%%%%%%%%%%%%%%%%%%%
% The proposed model provides intervention strategies for modifiable risk factors to offset the effect of MCC progression in patients.
%%%%%%%%%%%%%%%%%%%%%%%%%%%%%%%%%%%%%%%%
The proposed method is tested under different scenarios (e.g., age group, the prior existence of MCC), demonstrating the effective intervention strategies for improving the lifestyle behavioral risk factors to offset MCC evolution.

%\noindent\texttt{\textbackslash begin{abstract}} \dots 
%\texttt{\textbackslash end{abstract}} and
%verb+\begin{keyword}+ \verb+...+ \verb+\end{keyword}+ 

\end{abstract}

% \includegraphics{figs/grabs.pdf}

%\item Research highlights item 1
%item Research highlights item 2
%\item Research highlights item 3

% Keywords
\begin{keywords}
Model Predictive Control \sep Functional Continuous Time Bayesian Network
\sep Multiple Chronic Conditions\sep Group Regularization\sep Structure Learning
\end{keywords}

\maketitle

\section{Introduction}

\label{Section: Introduction}
The emergence and progression of multiple chronic conditions (MCC) forms a complex stochastic network of interconnected medical conditions with dependencies regulated by patients modifiable lifestyle behavioral factors as well as non-modifiable demographic risk factors \cite{faruqui2021functional}.
toward that direction, a healthy lifestyle positively impacts the connections, imparting a longer life expectancy free of major chronic conditions (e.g. diabetes, cardiovascular disease) \cite{li2020healthy}, improved weight and lipid level control \cite{kuna2013long}, and reversion or slowing of cognitive decline \cite{shimada2019reversible, kivipelto2018lifestyle, blumenthal2019lifestyle}. This might be because, a healthy lifestyle may induce epigenetic modification to regulate patho-physiologic cellular processes, e.g. protein expression and RNA coding genes, cell differentiation, embryo-genesis, DNA methylation, and repair regulation \cite{stylianou2019epigenetics}. It also may alter the oxidative balance and the dynamics of chronic inflammation, which are highly related to chronic conditions \cite{carraro2018physical, liguori2018oxidative, hussain2016oxidative, franceschi2014chronic}.

% While more than a quarter of all americans are estimated to have multiple chronic conditions (MCC), being proactive about managing MCC can certainly alleviate some of these burden.
%%%%%%%%%%%%%%%%%%%%%%%%%%%%%%%%%%%%%%%%%%%%%%%%%%%%%%%%%%%%%%%%%%
%%%%%%%%%%%%%%%%%%%%%%%%%%%%%%%%%%%%%%%%%%%%%%%%%%%%%%%%%%%%%%

One of the major aspects of MCC that has been extensively studied in the literature is the impact of different risk factors on MCC network. 
%%%%%%%%%%%%%%%%%%%%%
\cite{alaeddini2017mining} Proposed a latent regression Markov clustering to identify major transitions between chronic conditions in MCC networks, and validated the proposed method on a large dataset from the department of veteran affairs. 
%%%%%%%%%%%%%%%%%%%%%
\cite{faruqui_mining_2018} developed an unsupervised multi-level temporal Bayesian network to provide a compact representation of the relationship among emergence of multiple chronic conditions and patient level risk factors over time. They also used a longest path algorithm to identify the most likely sequence of comorbidities emerging from and/or leading to specific chronic conditions. 
%%%%%%%%%%%%%%%%%%%%%
\cite{faruqui2021functional} proposed a continuous time Bayesian network with conditional dependencies represented as regularized Poisson regressions to model the impact of exogenous variables on the conditional intensities of the MCC network of five chronic conditions.
%%%%%%%%%%%%%%%%%%%%%
\cite{faruqui2021dynamic} proposed dynamic continuous time Bayesian networks to formulate the dynamic effect of patients’ modifiable lifestyle behaviours and their interaction with non-modifiable  socio-demographics and preexisting conditions on the emergence of MCC. They considered the parameters of the conditional dependencies of MCC as a nonlinear state-space model and develops an extended Kalman filter to capture the dynamics of the modifiable risk factors on the MCC network evolution. 
%%%%%%%%%%%%%%%%%%%%%%%%%%%%%%%%%%%%%%%%%%
\cite{Isvoranu2021ExtendedNA} proposed a weighted network built using Ising model and eLasso technique conjointly with the extended Bayesian Information Criterion. They used this network to analyze the relationship between psycho-pathologies and potential pathways for developing MCC. Their finding indicate gender as a risk factor affects the pattern of comorbidity among patients with hypertension (with heart and multi-organ failure), cerebral vascular disease and anxiety. 
%, building a weighted network using multiple measures from an administrative data, and fitting the data using an Isign model and eLasso technique,   
%%%%%%%%%%%%%%%%%%%%%%%%%%%%%%%%%%%%%%%%%%%
To analyze the co-occurance relationship between pairwise comorbodities, Zhou et al \cite{Zhou2022PhenotypicDN} built a phenotypic disease network. They further applied community detection algorithm to identify cluster of co-occuring conditions within the comorbidity network. 
% Faruqui et al. \cite{faruqui_mining_2018, faruqui2021dynamic} showed in two separate studies the effect of risk factors (both modifiable and non-modifiable) in MCC progression. In their study \cite{faruqui2021dynamic} they showed how possible detection of early onset of MCC can help physicians better manage patients healthcare needs.
%%%%%%%%%%%%%%%%%%%%%%%%
Lippa et al. \cite{lippa2015deployment} used factor analysis to identify patterns of comorbidity in Post-9/11 deployed service members and veterans who participated in their clinical interviews. They identified four clinically relevant psychiatric and behavioral factors, including deployment trauma factor, somatic factor, anxiety factor, and substance abuse factor, that account for 76.9\% of the variance in the data. 
%%%%%%%%%%%%%%%%%%%%%%%%
Cai et al. \cite{cai_analysis_2015} used Bayesian network using a tree-augmented naive bayes algorithms to identify the relationships between factors influencing hepatocellular carcinoma after hepatectomy. 

%%%%%%%%%%%%%%%%%%%%%%%%
% Several studies have also covered patient support and complications \cite{wyatt_out_2014, beadles_medical_2015}; and assessment and prediction \cite{pugh_complex_2014, miotto_deep_2016, alaeddini_mining_2017, faruqui_mining_2018}.
%%%%%%%%%%%%%%%%%%%%%%%%%%%%%%%%%
% Along with predictive methods for modeling the impact of patient level risk factors on the evolution of MCC, 
Prescriptive methods have also been increasingly used in recent years for control and management of MCC network evolution based on patients' modifiable risk factors. Methods based on control theory \cite{baumann2012helping} are among the most common approaches that have been used to manage chronic conditions \cite{zurakowski2006model, nanda2007optimal}. However, due to the complexity of medical decision making, most of these decision support tools focus on only a single chronic condition. Markov decision processes (MDP)\cite{alagoz2010markov} and partially observable MDP \cite{hauskrecht2000planning} have also been commonly used for screening and treatment of chronic diseases \cite{steimle2017markov, magni2000deciding}. However, MDP models do not scale up well to large state spaces and hence have been mostly used for a very limited range of actions, i.e., identifying the next checkup time. Finally, reinforcement learning methods \cite{li2017deep} have been increasingly used in recent years for designing clinical trials and dynamic treatment regimes in chronic conditions \cite{pham2016deepcare, zhang2019near}. However, reinforcement learning methods are usually computationally extensive and do not provide explainability. 
On the other hand, model predictive controls (MPC) are best suited for this purpose. In conjuncture with non-linear state-space models like FCTBN\cite{faruqui2021functional,faruqui2021dynamic} they can represent the complex relationship of MCC, and explain the impact of risk-factors on the MCC emergence, progression and generates intervention strategies \cite{mayne2000constrained, kothare1996robust}. In recent years MPC have been deployed to study COVID-19 spreading dynamics and find mitigating strategies to its rapid spreading while minimizing its adverse impact on society and economy\cite{peni2021convex, mayne2014model}.

%%%%%%%%%%%%%%%%%%%%%%%%%%%%%%%%%%%%%%%%%%%%%%%%%%%%%%%%%%%%%%
%%%%%%%%%%%%%%%%%%%%%%%%%%%%%%%%%%%%%%%%%%%%%%%%%%%%%%%%%%%%%%
\begin{figure*}[!t]
    \centering
    {
    \subfloat[]{\label{Figure:Overall_Schema}\includegraphics[scale = 0.28]{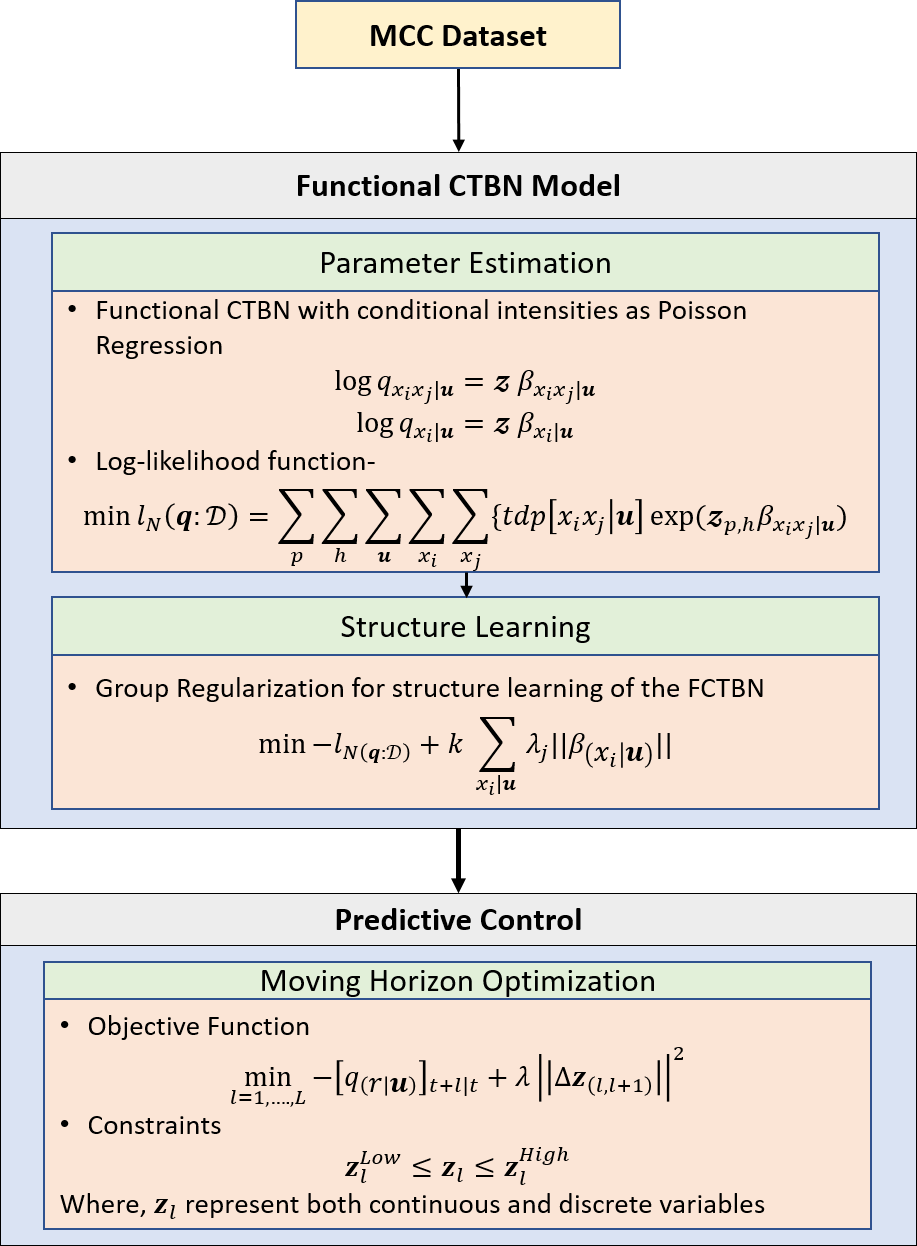}}\quad%
    \subfloat[]{\label{Figure:Overall_Intervention_Schema}\includegraphics[scale = 0.28]{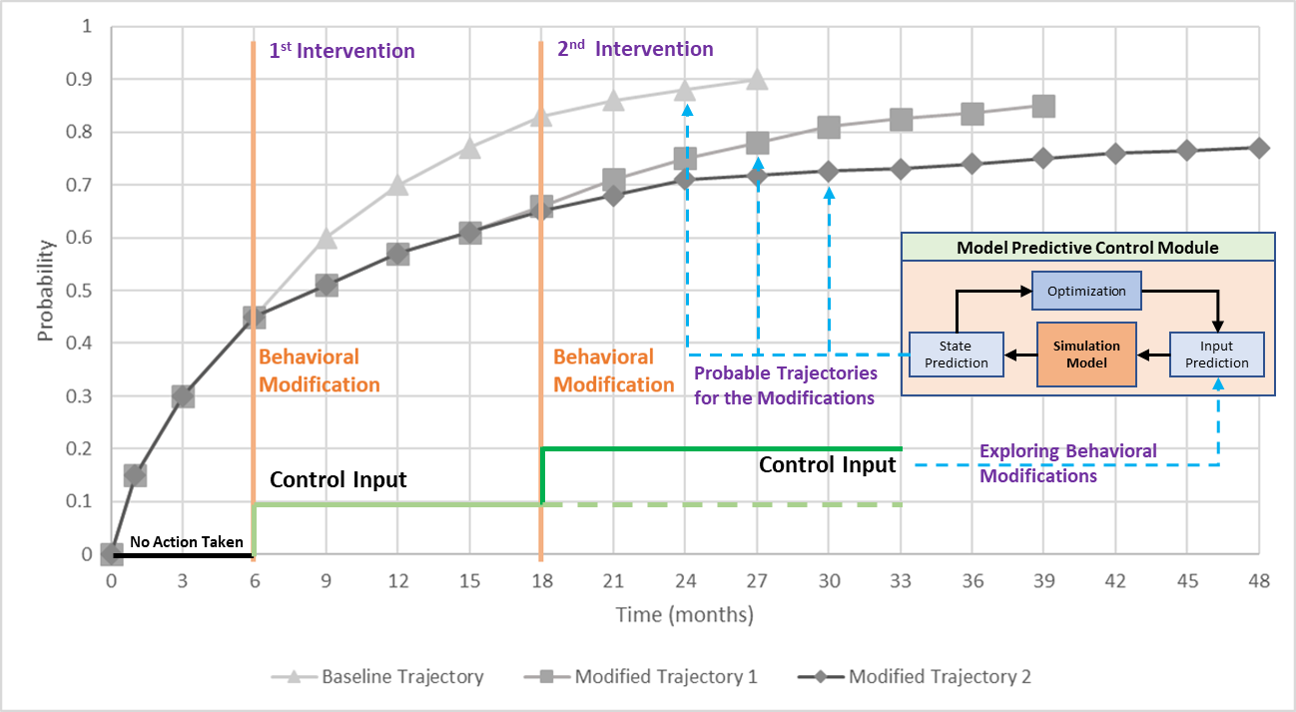}}\\%
    }
    \caption{(a) Overall schema of the proposed approach for exploring intervention strategies. The proposed scheme has two major components: (1) A functional CTBN (FCTBN) to take into account the impact of the patients' (static) risk factors on the MCC emergence and progression, (it can also be replaced with a dynamic FCTBN (D-FCTB) for dynamic prediction whenever needed), and (2) The model predictive control (MPC). (b) Illustration of how the MPC model recommends behavioral changes by making multi-step-ahead prediction.}
    \label{Figure:Overall_MPC}
\end{figure*}
%%%%%%%%%%%%%%%%%%%%%%%%%%%%%%%%%%%%%%%%%%%%%%%%%%%%%%%%%%%%%%
%%%%%%%%%%%%%%%%%%%%%%%%%%%%%%%%%%%%%%%%%%%%%%%%%%%%%%%%%%%%%%

In this study, we integrate model predictive control (MPC) with functional continuous time Bayesian networks (FCTBN) to create an online estimation and decision making procedures for MCC networks to (1) examine the impact of various lifestyle behavioral changes on the emergence trajectories of MCC, and (2) suggest personalized lifestyle behavioral changes to minimize the risk of progression of chronic conditions. 
% We hypothesize that specific changes to  modifiable lifestyle behaviours will minimize the risk of new conditions emergence or delay their onset.
We first represent the complex stochastic relationship between MCC as a functional continuous time Bayesian network (FCTBN) \cite{faruqui2021functional} to take into account the impact of the patients' risk factors on the MCC emergence and progression (Figure \ref{Figure:Overall_Schema}: first component - functional CTBN Model).
We also derive a mini-batch gradient descent algorithm for updating the FCTBN conditional intensities, which are based on Poisson regression, for new longitudinal patient observations.
%%%%%%%%%%%%%%%%%%%%%%%%
Next, we integrate the FCTBN with an MPC to generate intervention strategies based on modifiable lifestyle factors. This is done by formulating conditional dependencies of FCTBN as a non-linear state-space model (Figure \ref{Figure:Overall_Schema}: second component - Predictive Control). The objective function is to minimize the conditional intensity of transition for emergence of new MCC. We also develop constraints for controlling the suggested changes in the lifestyle behaviours per medical requirement. As the constraints can be adjusted for the individual patients depending on their pre-existing conditions and static  socio-demographic risk factors, they make the proposed model more personalizable.

% Thus, we propose to develop a model predictive control (MPC) to look for lifestyle changes that provide the lowest predicted risk trajectory of MCC progression and minimize the onset of a new chronic condition in patients with MCC.
%%%%%%%%%%%%%%%%%%%%%%%%
The justification for using FCTBN is that, FCTBN: (1) provides closed form formulation for time to emergence of new chronic conditions based on changes in lifestyle behaviors, (2) generate multi-step-ahead predictions and trajectories of MCC evolution over time, (3) provides uncertainties of the predictions, (4) can be well integrated with optimization algorithms.
% identifying the short- and long-term impact of changes in lifestyle behaviours requires multi-step-ahead predictions, which can be efficiently provided by the FCTBN trajectory analysis. 
Also, the justification for using MPC is because MPC: (1) provides significant computational efficiency, (2) offers closed-form formulation for optimal time of behavioral change, (3) allows for sensitivity analysis, and (4) provides the trajectory of MCC and self-management strategies.
Meanwhile, given the modular structure of the proposed method, both the FCTBN and/or the MPC model can be replaced with other relevant prediction and decision making models, i.e, D-FCTBN \cite{faruqui2021dynamic}, and reinforcement learning \cite{yu2021reinforcement}. 

%%%%%%%%%%%%%%%%%%%%%%%%
Figure \ref{Figure:Overall_Intervention_Schema} shows a schematic diagram of the overall process, where, at each stage of patient monitoring, i.e follow up visits, possible trajectories of existing/non-existing MCC are being generated with respect to the patient's lifestyle behavioral factors. Simultaneously, the model generates the possible modifiable lifestyle factors to possibly minimize the MCC trajectory. The procedure continues recursively at each stage to perform the same operation.
%%%%%%%%%%%%%%%%%%%%%%%%

The remained of the manuscript is structured as follows. Section \ref{Section: Relevant_Background} presents the preliminaries and background for CTBN and MPC. Section \ref{Subsection:Model_Predictive_Control} details the objective function and the constraints of MPC. Section \ref{Section:Result_Discussion} presents the study population and simulated case scenario for MCC patients. Finally section \ref{Section: Conclusion} provides the concluding remarks.

All the above packages are part of any
standard \LaTeX{} installation.
Therefore, the users need not be
bothered about downloading any extra packages.

\section{Relevant Background}
\label{Section: Relevant_Background}
In this section, we review some of the major components of the proposed approach including CTBN for modeling MCC evolution as a finite-state continuous-time conditional Markov process over a factored state \cite{nodelman_continuous_2012}, 
CTBN parameter learning \cite{nodelman2012expectation, nodelman_learning_2012}, and MPC for online control \cite{zurakowski2006model}.
% including the CTBN for modeling MCC evolution as a finite-state continuous-time conditional Markov process over a factored state \cite{nodelman_continuous_2012, nodelman2012expectation, nodelman_learning_2012}, and functional CTBN (FCTBN) \cite{faruqui2021functional} for extending CTBN edges based on Poisson regression of some exogenous risk factors.
%% ############################################################################
% \vspace{-1em}
\subsection{Continuous Time Bayesian Network (CTBN)}
\label{Subsection:Continuous_Time_Bayesian_Network}

\subsubsection{CTBN Components}
\label{Subsubection:CTBN1}
Continuous time Bayesian networks (CTBNs) are Bayesian networks that models time explicitly by defining a graphical structure over continuous time Markov processes \cite{nodelman_continuous_2012}. 
Let $X = \{x_1,x_2,....,x_n\}$ denotes the state space of a set of random variables with 
discrete states $x_i=\{1,...,l\}$, such as MCC like diabetes (DI), Obesity (OB), cognitive impairment (CI), hyperlipidemia (HL), and hypertension (HP).
A CTBN consists of a set of conditional intensity matrices (CIM) under a given graph structure \cite{nodelman_continuous_2012, norris_markov_1998}. The components of a CTBN are - 

% ################################################
\begin{enumerate}
  \item An initial distribution $(P_x^0)$, which formulates the structure of the (conditional) relationship among the random variables and is specified as a Bayesian network, where each edge $x_i \to x_j$ on the network implies the impact of the parent condition $x_i$ on the child condition $x_j$.
  \item A state transition model $(Q_{X_i|\textbf{u}})$, which describes the transient behavior of each variable $x_i\in X$ given the state of parent variables $\textbf{u}$, and is specified based on CIMs -
\end{enumerate}
% \noindent $(Q_{X|\textbf{u}})$ can be written as-
\[
\textbf{Q}_{X|\textbf{u}} = \begin{bmatrix} 
    -q_{x_1|\textbf{u}}    & q_{x_1x_2|\textbf{u}}    & \dots & q_{x_1x_n|\textbf{u}} \\
    q_{x_2x_1|\textbf{u}}  & -q_{x_2|\textbf{u}}      & \dots & q_{x_2x_n|\textbf{u}} \\
    \vdots      & \vdots        & \ddots & \vdots\\
    q_{x_nx_1|\textbf{u}}  & q_{x_nx_2|\textbf{u}}    & \dots & -q_{x_n|\textbf{u}}
    \end{bmatrix}
\]
\noindent where $q_{x_ix_j|\textbf{u}}$ represents the intensity of the transition from state $x_i$ to state $x_j$ given a parent set of node $\textbf{u}$, and $q_{x_i} = \sum_{j \neq i} q_{x_ix_j}$.  
% ################################################
Conditioning the transitions on parent conditions sparsifies the intensity matrix considerably, which is especially helpful for modeling large state spaces. When no parent variable is present, the CIM will be the same as the classic intensity matrix.

The probability density function ($f$) and the probability distribution function ($F$) for staying at the same state (say, $x_i$), which is exponentially distributed with parameter $q_{x_i}$, are calculated as-

%%%%%%%%%%%%%%%%%%%%%%%%%%%%%%%%%%%%%%%%
\begin{align}
\label{Equation:Probability density}
    f(q_x,t) &= q_{x_i} exp(-q_{x_i}t),&{t \geq 0}\\
    F(q_x,t) &= 1 - exp(-q_{x_i}t),    &{t \geq 0}
\end{align}
%%%%%%%%%%%%%%%%%%%%%%%%%%%%%%%%%%%%%%%%

\subsubsection{CTBN Parameter Estimation}
\label{Subsubsection:Parameter_Estimation1}
Given a dataset $\mathcal{D} = \{\tau_{h=1}, \tau_{h=2},....,\tau_{h=H}\}$ of $H$ observed transitions, where $\tau_h$ represents the time at which the $h^{th}$ transition has occurred, and $\mathcal{G}$ is a Bayesian network defining the structure of the (conditional) relationship among variables, we can use maximum likelihood estimation (MLE) (equation \eqref{Equation:Likelihood_Function}) to estimate parameters of the as defined in Nodelman et al \cite{nodelman_continuous_2012, nodelman_learning_2012}-

%%%%%%%%%%%%%%%%%%%%%%%%%
\begin{align}
\label{Equation:Likelihood_Function}
    \begin{split}
    L_x(q_{x|\textbf{u}}:\mathcal{D}) &= \prod_{\textbf{u}} \prod_x q_{x|\textbf{u}}^{M{[x|\textbf{u}]}} exp(-q_{x|\textbf{u}}T{[x|\textbf{u}]})
    \end{split}
\end{align}
%%%%%%%%%%%%%%%%%%%%%%%%%

\noindent where, $T[x|\textbf{u}]$ is the total time $X$ spends in the same state $x$, and $M{[x|\textbf{u}]}$ the total number of time $X$ transits out of state $x$ given, $x = x'$ .The log-likelihood function can be then written as$-$

%%%%%%%%%%%%%%%%%%%%%%%%%
\begin{align}
\label{Equation:Log-Likelihood_Function}
    \begin{split}
    l_x(q_{x|\textbf{u}}:\mathcal{D}) &= \sum_{\textbf{u}} \sum_x M[x|\textbf{u}]\hspace{1pt} ln(q_{x|\textbf{u}}) - q_{x|\textbf{u}}\hspace{1pt} T[x|\textbf{u}]
    \end{split}
\end{align}
%%%%%%%%%%%%%%%%%%%%%%%%%
Maximizing equation \eqref{Equation:Log-Likelihood_Function}, provides the maximum likelihood estimate of the paramters of the FCTBN. 

\subsection{Model Predictive Control}
\label{Subsection:Model_Predictive_Control_}
MPC is an online iterative control algorithm that solves a constrained optimization problem to determine the control parameters at each time step of a system or at decided point of interest. In the MPC approach, the current control action is computed on-line rather than using a pre-computed, off-line model. 
MPCs are used in majority of existing multivariable control applications \cite{zurakowski2006model}. This is due the nature of its straightforward formulation, based on well understood concepts; ease of maintenance while changing system specs and shorted development time.

\section{Proposed Methodology}
\subsection{Functional CTBN (FCTBN)}
\label{Subsubsection:Functional_CTBN}

\subsubsection{FCTBN with Conditional Intensities as Poisson Regression}
\label{Subsubection:FCTN_PoissonRegression}
In reality, the progression of state variables, such as chronic conditions, not only depends on the state of their parents, such as preexisting chronic conditions but some exogenous variables, such as patient level risk factors like age, gender, etc. Using Poisson regression to represent the impact of exogenous variables on the conditional dependencies, the rate of transition between any pair of MCC states can be derived as \cite{faruqui2021functional}-
%%%%%%%%%%%%%%%%%%%%%%%%%%%%%%%%%%%%%%%%
\begin{subequations}
    \begin{align}
            \label{Equation:CTBN_Poisson_1}
                % \log{q_{x_ix_j|\mathit{u}}}=\beta_{0_{x_ix_j|\mathit{u}}}+\ldots+{\beta_{m_{x_ix_j|\mathit{u}}}z}_m=\mathit{z}\mathit{\beta}_{x_ix_j|\mathit{u}}\\
        \log {q}_{x_i,x_j|\textbf{u}} &= \beta_{0_{x_i,x_j|\textbf{u}}} + \beta_{1_{x_i,x_j|\textbf{u}}} + ... ... + z_m \beta_{m_{x_i,x_j|\textbf{u}}} \\
                        &= \boldsymbol{z} \boldsymbol{\beta}_{\textbf{x}_i,\textbf{x}_j|\textbf{u}}
    \end{align}
\end{subequations}

%%%%%%%%%%%%%%%%%%%%%%%%%%%%%%%%%%%%%%%%
\noindent where, $\textbf{z}=\{z_1,z_2,...,z_m\}$ is the set of exogenous variables (e.g. patient-level risk factors such as age, gender, education, marital status, etc.), and $\beta_{k_{x_i|\textbf{u}}}=\sum_{j\neq i}\beta_{k_{x_ix_j|\textbf{u}}}, k=0,\ldots,m$ is the set of coefficients (parameters) associated with the exogenous variables. 
% Given the conditional intensity matrix (CIM) property of $\log{q_{x_i|\mathit{u}}}=-\log{q_{x_ix_j|\mathit{u}}}$, 
Also, the rate of staying in the same state is modeled as-
%%%%%%%%%%%%%%%%%%%%%%%%%%%%%%%%%%%%%%%%
\begin{subequations}
    \begin{align}
        \label{Equation:CTBN_Poisson_2}
        \log q_{x_i|\textbf{u}} &= \beta_{0_{x_i|\textbf{u}}} + z_1 \beta_{1_{x_i|\textbf{u}}} + ... ... + z_m \beta_{m,{x_i|\textbf{u}}} \\
                        &= \boldsymbol{z} \boldsymbol{\beta}_{\textbf{x}_i|\textbf{u}}
    \end{align}
\end{subequations}
%%%%%%%%%%%%%%%%%%%%%%%%%%%%%%%%%%%%%%%%

When the state space of the random variables is binary, as in our case study on MCC transitions, where MCC states include having/not having each of the conditions, the conditional intensities in $\textbf{Q}_{x_i |\textbf{u}_i}$, can be estimated just using Equation  \ref{Equation:CTBN_Poisson_2} because for Markov processes with binary states $ q_{x_i |\textbf{u}} = - \sum_{j\neq i} q_{(x_i x_j |\textbf{u})}$. 
This feature considerably simplifies the estimation of the functional CTBN conditional intensity matrix based on Poisson regression.

%% ############################################################################
\subsubsection{Parameter Estimation}
\label{Subsubsection:Parameter_Estimation}
Having the dataset $D=\left\{\tau_{\left(p=1,h=1\right)},\ldots,\tau_{\left(P,H\right)}\right\}$ of MCC trajectories, where $\tau_{\left(p,h\right)}$ represents the time at which the $h^{th}$ (MCC) transition of  the $p^{th}$ patient has occurred, we use maximum likelihood estimation to estimate parameters of the proposed FCTBN. The likelihood of $D$ can be decomposed as the product of the likelihood for individual transitions. Let $d=\langle \textbf{z},\textbf{u},x_i|\textbf{u},t_d,x_j|\textbf{u} \rangle$ be the transition of patient $p$ with risk factors $\mathit{z}$ and existing conditions $\textbf{u}$, who made the transition to state $x_{j|\textbf{u}}$ after spending the amount of time $t_d=\tau_{\left(p,h\right)}-\tau_{\left(p,h-1\right)}$ in state $x_{i|\textbf{u}}$. By multiplying the likelihoods of all conditional transitions during the entire trajectory for all patients $p=1,\ldots,P$, and taking the log, we obtain the overall log$-$likelihood function as$-$ 
\begin{align}
\label{Equation:FCTBNLikelihood_Function}
    \begin{split}
    l_N\left(\textbf{q}:\mathcal{D}\right)\
    &=\sum_{p}\sum_{h}\sum_{\textbf{u}}\sum_{x_j}\sum_{x_i} \\
    & \left\{{t_d}_p\left[x_ix_j|\textbf{u}\right]\exp{\left(\boldsymbol{z}_{p,h}\boldsymbol{\beta}_{x_ix_j|\textbf{u}}\right)}\right\}
    \end{split}
\end{align}

which is a convex function and can be maximized efficiently using a convex optimization algorithm such as Newton$-$Raphson to estimate parameters $\mathit{\beta}_{x_i|\textbf{u}}$. 

\subsubsection{Group regularization for structure learning of the FCTBN}
\label{Subsubsection:Group regularization}
The parameter estimation approach presented above requires the parent set of each condition to be known, which is equivalent to knowing the Bayesian network structure. Given that FCTBN has a special structure based on a conditional intensity matrix that allows for cycles, group regularization can be used to penalize groups of parameters pertaining to each specific conditional transition (each edge) \cite{faruqui2021functional} as-
\begin{equation}
    \label{Equation:CTBN_Minimization_1}
    \centering
    \min -l_N(\textbf{q}:\mathcal{D}) + k \sum_{x_{i}|\textbf{u}}\lambda_j\|\boldsymbol{\beta}_{x_{i}|\textbf{u}}\|^2
\end{equation}

\noindent where, $\|\boldsymbol{\beta}_{x_{i}|\textbf{u}}\| = \sqrt{\sum_{\textbf{u}} \sum_{x_{i}} (\boldsymbol{\beta}_{x_{i|\textbf{u}}}.\boldsymbol{\beta}^T_{x_{i|\textbf{u}} )}}$ is the $L_1$-norm of the group of parameters associated with each conditional transition. $k$ is the groups size which is based on the number of coefficients in the Poisson regression for each conditional intensity. $\lambda_j = \lambda \|\Tilde{\boldsymbol{\beta}_j}\|^{-1}$ is the tuning parameters (of the adaptive group regularization) that control the amount of shrinkage, where $\lambda$ is inversely weighted based on the unpenalized estimated value of the regression coefficients $\Tilde{\boldsymbol{\beta}_j}$ \cite{wang_note_2008}.

\cite{faruqui2021functional} proposed to a Lasso penalty for the group regularization based structure learning of the CTBN, namely $|\boldsymbol{\beta}_{x_{i}|\textbf{u}}\|$. Here, we propose to use a ridge penalty instead of Lasso, because: (1) even though ridge penalty would result in less sparsity, if applied appropriately along with early termination criteria or regularization path analysis, it can provide the same level of sparsity as Lasso, while providing significant computational advantage, and (2) ridge regularization provides a straightforward way for incorporating online updation of the model parameters based on new patients' observations using mini-batch gradient descents as-

\begin{equation}
    \label{Equation:gradient descent}
    \begin{split}
    \boldsymbol{\beta}_{x_{i}|\textbf{u}}^{(New)}=\boldsymbol{\beta}_{x_{i}|\textbf{u}}^{(Old)} - \eta
    &\sum_{p'\in p}\sum_{h'\in h}\sum_{\textbf{u}}\sum_{x_j}\sum_{x_i} \\
    & \boldsymbol{z}_{p,h} \left\{{t_d}_p\left[x_ix_j|\textbf{u}\right]\exp{\left(\boldsymbol{z}_{p,h}\boldsymbol{\beta}_{x_ix_j|\textbf{u}}\right)}\right\}
    \end{split}
\end{equation}

% \begin{landscape}
    \begin{figure*}[!t]
        \centering
        \subfloat[]{\label{Figure_3_MPC}\includegraphics[scale = 0.15]{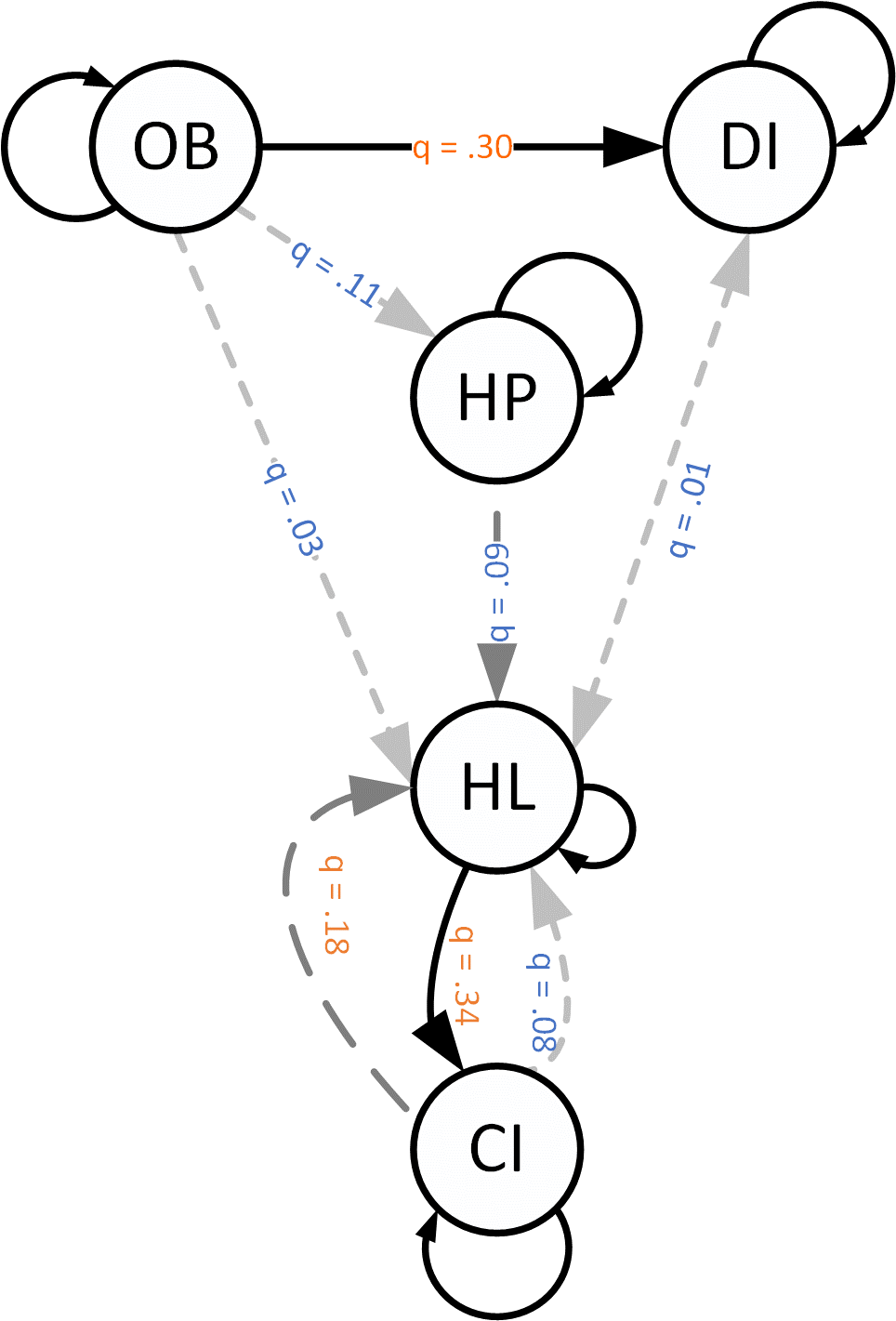}}\quad%
        \subfloat[]{\label{Figure_4_MPC}\includegraphics[scale = 0.21]{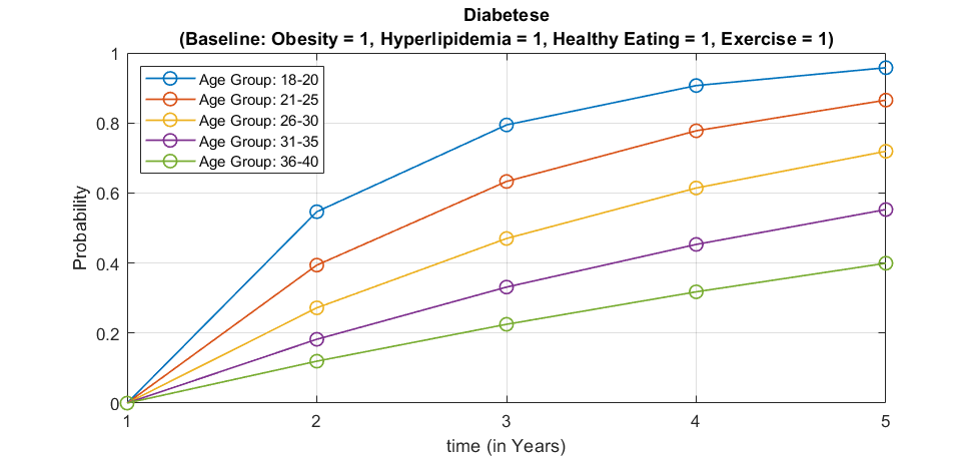}}\\%
        \caption{(a) Network Structure of the F-CTBN model for 5 MCC, including
        diabetes (DI), Obesity (OB), cognitive impairment (CI), hyperlipidemia (HL), and hypertension (HP). The self-loops represent keeping the conditions over time. (b) The risk trajectory of developing diabetes with the prior conditions of Obesity and hyperlipidemia when eating healthy and exercise for different age groups over five years. }%
        \label{Figure:Illustration}
    \end{figure*}
% \end{landscape}

\subsection{Model Predictive Control}
\label{Subsection:Model_Predictive_Control}
\subsubsection{Moving-horizon optimization of behavioral changes}
\label{Subsubsection:Moving-Horizon}
The MPC optimization model recommends changes in behaviours that make the multi-step-ahead prediction of MCC as close as possible to a reference (ideal) trajectory of MCC in which no progression or emergence of new MCC occurs for the next several follow-up periods. The proposed optimization model is a constrained moving horizon convex optimization problem that identifies the sequence of behavioral changes at each follow-up visits of the CCHC study, which efficiently maximizes the probability of following the reference trajectory over the prediction horizon, i.e. \[\max_{l = 1, ..., L} P_{[r|\textbf{u}]_{t+l|t}}(\Delta t = 1)\]

\noindent where $\Delta t = (t+1) - t = 1$ represents the time between the current and the next immediate follow-up visit. For optimization convenience, we rewrite the model as a penalized negative log-likelihood problem, which after some algebraic simplification result in Eq. \ref{Equation: Obj_func_MPC}, 

\begin{equation}
    \min_{l=1, ..., L} - q_{(r|\textbf{u})}{}_{t+l|t} + \lambda || \Delta \textbf{z}_{(l, l+1)}||^2
    \label{Equation: Obj_func_MPC}
\end{equation}

\noindent and constraints in \ref{Equation: Constr_MPC}. 
\begin{equation}
    \textbf{z}^{Low}_l \leq \textbf{z}_l \leq \textbf{z}^{High}_l 
    \label{Equation: Constr_MPC}
\end{equation}

\noindent where, $z_l$ is the (suggested) behavior change for the period $t+l$ (in the future), $\Delta \textbf{z}_{(l,l+1)}$ is the change in the behavioral factor since the last follow-up visit, $\lambda$ is a weighting parameter to ensure there will not be a drastic change in suggestions from one period to another, i.e., sudden weight loss, and inequalities represent constraints on suggested behaviours to maintain them within range, i.e., keeping food intake within a safe range. MPC solves the optimization model in Eq. \ref{Equation: Obj_func_MPC} every time a new observation of MCC is available, or a medical practitioner decides to investigate, but only applies the behavioral change for the next period $t+1$. This strategy, which is known as the receding horizon, dynamically updates the suggested behavioral change and the predicted trajectory of MCC after each new observation. This provides an effective tool for the patients (and the practitioners) to evaluate the short- and long-term impact of different self-management strategies (See figure \ref{Figure:Overall_Intervention_Schema}). 

\section{Result and Discussion}
\label{Section:Result_Discussion}
\subsection{Study Population and Demographics}
\label{Subsection:Study_Population}
\begin{figure}[t]
    \centering
    \includegraphics[width=.45\textwidth]{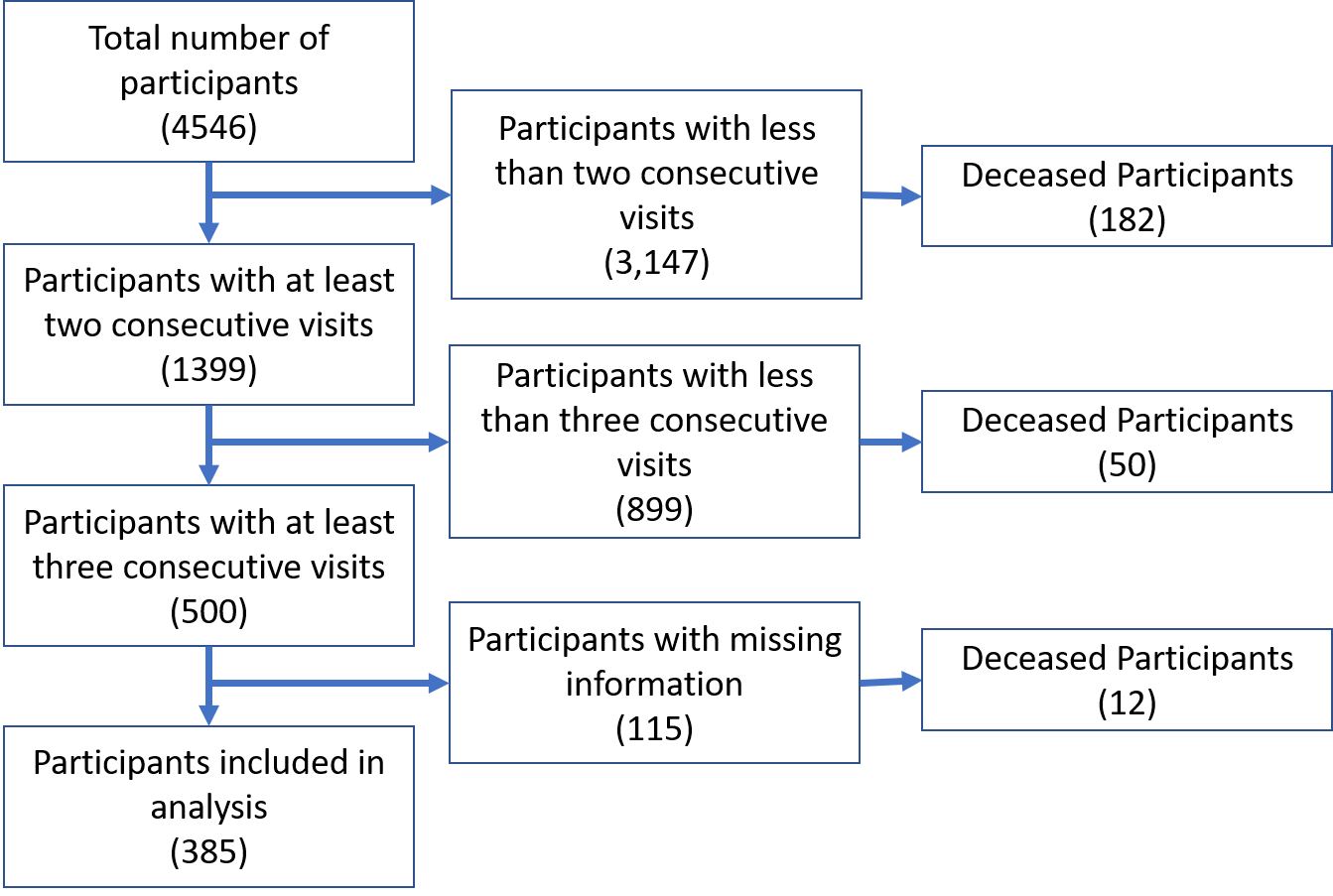}
    \caption{Flow diagram of sample selection and the final number of patients included in the analysis.}
    \label{Figure:Study_Population_Selection}
\end{figure}

The case study is based on the Cameron County Hispanic Cohort (CCHC) dataset for comorbidity analyses. The CCHC is a cohort study comprised of mainly Mexican Americans (98\% of cohort) randomly recruited from a population with severe health disparities on the Texas-Mexico border and started in 2004. 
% The CCHC is funded by the National Center for Minority Health Disparities.
The CCHC is employing a rolling recruitment strategy and currently numbers 4,546 adults. Inclusion criteria: (1) participating in the study between 2004 and 2020, (2) having at least three 5-year follow up visits during that period. 385 patients met these criteria, which include the dataset of our study (see Figure \ref{Figure:Study_Population_Selection}). The survey includes participants’ socio-demographic factors (age, gender, education, marital status,  etc.) and lifestyle behavioral factors (diet, exercise, tobacco use, alcohol use, etc.). 

%%%%%%%%%%%%%%%%%%%%%%%%%%%% CASE: 1 %%%%%%%%%%%%%%%%%%%%%%%%%%%%%%%%
% \begin{landscape}
    \begin{figure*}[!t]
        \centering
        \subfloat[]{\label{Figure_5a_MPC}\includegraphics[scale = 0.20]{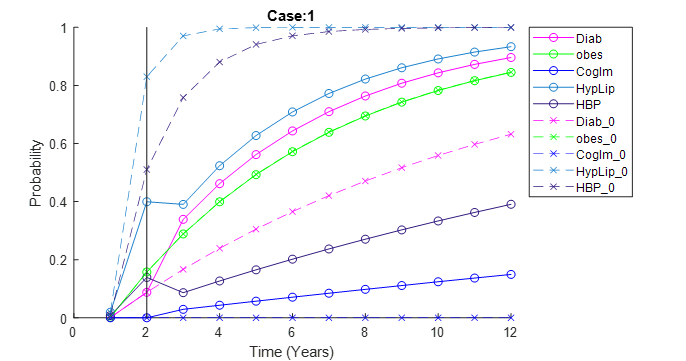}}\quad
        \subfloat[]{\label{Figure_5b_MPC}\includegraphics[scale = 0.20]{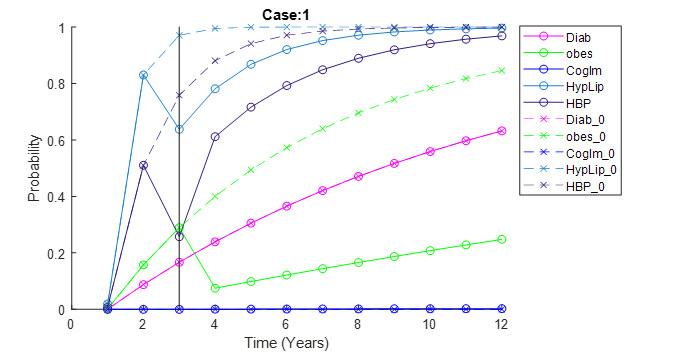}}\\
        \subfloat[]{\label{Figure_5c_MPC}\includegraphics[scale = 0.20]{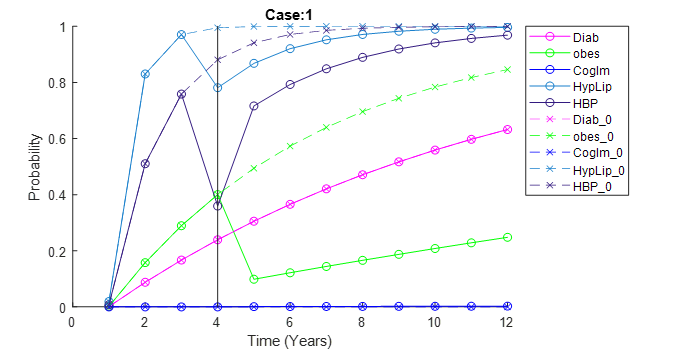}}\quad
        \subfloat[]{\label{Figure_5d_MPC}\includegraphics[scale = 0.20]{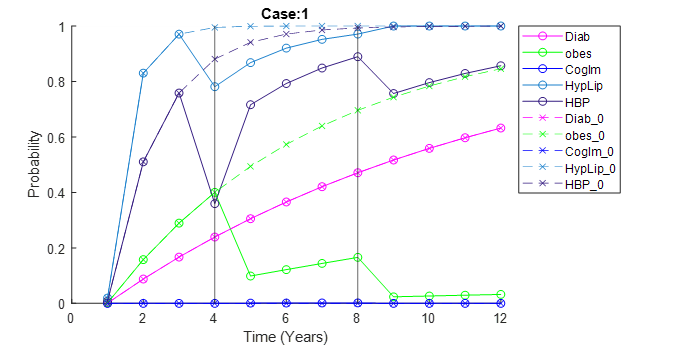}}
        \caption{Case Study 1: (a - c) Effect of single interventions at different time points in MCC evolution trajectories, (d) Effect of multiple interventions at different time points in MCC evolution trajectory. The vertical lines in the graphs shows the point of intervention (deployment of MPC algorithm) in time. The dashed line shows the probability of acquiring a condition without any intervention and solid line shows the probability of acquiring a condition with intervention.}%
        \label{Figure_5_MPC}
    \end{figure*}
% \end{landscape}
%%%%%%%%%%%%%%%%%%%%%%%%%%%% CASE: 1 %%%%%%%%%%%%%%%%%%%%%%%%%%%%%%%%

\subsection{Diagnosed Health Condition and Patient Associated Risk Factors}
\label{Subsection:Study_Population_2}
For this study, we considered some of the most common MCCs present in the Hispanic community, including diabetes, obesity, hypertension, hyperlipidemia, and mild cognitive impairment. The positive criteria (considering the condition to be active) for the conditions selected as below-
\begin{itemize}
    \item \textbf{Diabetes (DI)}: Fasting Glucose $>=$ 126 mg/dL, HbA1c $>=$ 6.5\%, or take diabetes medication \cite{association_diagnosis_2010}.
    \item \textbf{Obesity (OB)}: Body mass index (BMI, kg/m2) $>=$ 30 \cite{fruhbeck_abcd_2019}.
    \item \textbf{Hypertension (HP)}: Systolic blood pressure (BP) $>=$ 130mmHg, Diastolic BP $>=$ 80 mmHg, or take anti-hypertensive medication \cite{pk_2018}.
    \item \textbf{Hyperlipidemia (HL}: Total cholesterol $>$ 200 mg/dL, triglycerides $>=$ 150 mg/dLl, HDLC $<$ 40 mg/dL (for male)/ HDLC $<$ 50 mg/dL (for female), LDLC $>=$130 mg/dL, or take medication for hyperlipidemia \cite{grundy_scott_m_2018_2019}.
    \item \textbf{Mild Cognitive Impairment (CI)}: Mini-Mental State Score $<$ 23 (out of 30) \cite{wu_association_2018}.
\end{itemize}

For the risk factors, the dataset includes the participant's non-modifiable risk factors based on socio-demographic information (age, gender, education, marital status ) and modifiable risk factors based on lifestyle behavioral risk factors (diet, exercise, tobacco use, alcohol use). Diet and exercise are categorized according to the U.S. Healthy Eating Guideline, and U.S Physical Activity Guideline \cite{wu_fruit_2019}.

\subsection{Learning the State-Space Model}
Figure \ref{Figure_3_MPC} shows the illustration of the FCTBN for 5 MCC, including OB, DI, HL, HP, CI, and different types of connections, including self-loops and reinforcing connections, based on the preliminary analysis using the CCHC dataset using the FCTBN model. In the illustration, solid black arrows indicate strong conditional dependencies (connections), and grey dashed colors show weak connections. The values on the arrows show the rate of conditional intensities. The self-loops represent keeping the conditions over time \cite{faruqui2021functional}. Clinicians can utilize this model to inquire about different modifiable risk factors affecting the trajectory of MCC emergence over time and help guide patient and clinician decisions to delay the onset of new chronic conditions. For example, we found that the likelihood of having emerging diabetes diagnosis for Mexican American patients with the prior conditions of Obesity and hyperlipidemia has an average of 10\% decrease across all age groups at two years, and a 15\% decrease at 5 years (less for the younger groups) when we compared those who followed the U.S. healthy diet and physical activity guidelines to those who did not (see figure \ref{Figure_4_MPC}). Thus, a clinician can (if needed) further investigate the effects of lifestyle (modifiable) choices (food habits) over diabetes and can help patients by providing daily food guidelines.

%%%%%%%%%%%%%%%%%%%%%%%%%%%% CASE: 2 %%%%%%%%%%%%%%%%%%%%%%%%%%%%%%%%
% \begin{landscape}
    \begin{figure*}[!t]
        \centering
        \subfloat[]{\label{Figure_6a_MPC}\includegraphics[scale = 0.20]{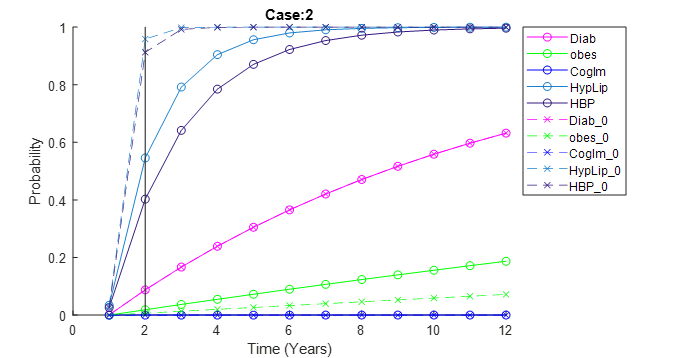}}
        \subfloat[]{\label{Figure_6b_MPC}\includegraphics[scale = 0.20]{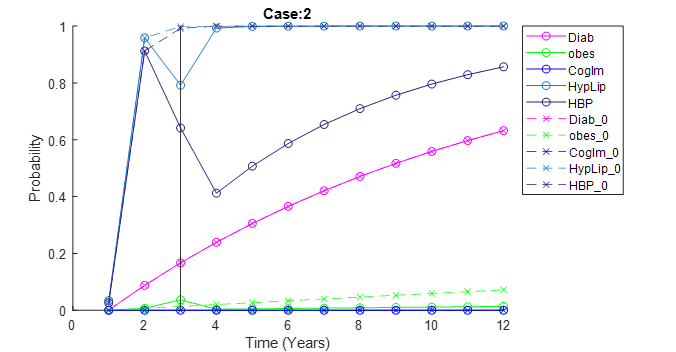}}\\
        \subfloat[]{\label{Figure_6c_MPC}\includegraphics[scale = 0.20]{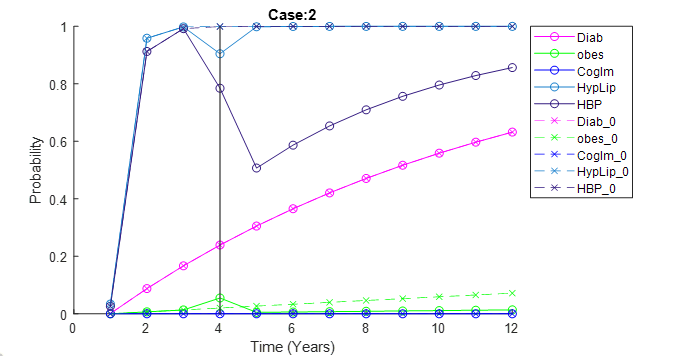}}
        \subfloat[]{\label{Figure_6d_MPC}\includegraphics[scale = 0.20]{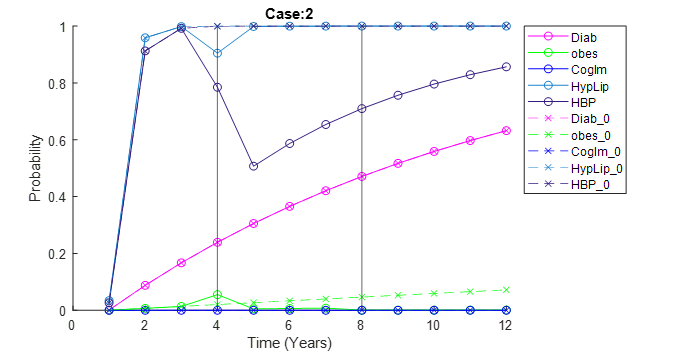}}
        \caption{Case Study 2: (a - c) Effect of single interventions at different time points in MCC evolution trajectories, (d) Effect of multiple interventions at different time points in MCC evolution trajectory. The vertical lines in the graphs shows the point of intervention (deployment of MPC algorithm) in time. The dashed line shows the probability of acquiring a condition without any intervention and solid line shows the probability of acquiring a condition with intervention.}%
        \label{Figure_6_MPC}
    \end{figure*}
% \end{landscape}
%%%%%%%%%%%%%%%%%%%%%%%%%%%% CASE: 2 %%%%%%%%%%%%%%%%%%%%%%%%%%%%%%%%

\subsection{Evaluation of Unsupervised Intervention Strategies}
Here, we demonstrate two case studies to examine the impact of various behavioral changes on the emerging trajectories of MCC in the CCHC dataset utilizing the proposed approach. We evaluate the mode for different preexisting conditions and age groups (exogenous variables). We further investigate the effect of intervention time (i.e., selecting a fixed time to intervene) and single versus multiple interventions. 
While the proposed MPC provides both the optimal timing as well as the optimal interventions for changing the lifestyle behaviours, In the case studies we only provide the results of suggested interventions at predetermined (fixed) times (year), for simplicity and better illustrations. Meanwhile, since MPC solves the optimization problem recursively for each and every time points, the extension of the results for optimal timing of interventions is trivial.

In the first case, we assumed a patient falls into the age group of $21 - 25$, doesn't follow a healthy eating habit, and doesn't perform any exercise. Figure \ref{Figure_5_MPC} shows the probability/trajectory of acquiring one or more conditions over time. The dashed line shows the probability change without any intervention, and the solid line shows the trajectory whenever there is an intervention. 

\noindent \textbf{Case (1-a): Single Intervention - Low Risk of Acquiring a New Condition:}
For figure \ref{Figure_5a_MPC}, Considering the clinician decides the intervention to be performed at year 2. The MPC algorithm performs an online optimization and suggests changing only the exercise (starting exercise) regiment can reduce the probability of hyperlipidemia, high blood pressure. At the same time, this affects the diabetes condition inversely (This can be due to the patient's eating habits). The probability of other conditions remained unchanged due to this change. 

\noindent \textbf{Case (1-b \& c): Single Intervention - High Risk of Acquiring a New Condition:}
For setup figure \ref{Figure_5b_MPC}, Assuming the clinician decides the intervention to be performed at year 3. This time, the proposed algorithm suggests that changing the exercise regiment (starting exercise) and food habit (start eating healthy) can reduce the probability of hyperlipidemia, high blood pressure, and obesity. The probability of obesity when following the MPC given suggestions is lower than the original estimation. This can be attributed to the lower possibility of problems related to obesity and hyperlipidemia when a healthy food habits is maintained and regular exercise is performed (defined as atleast walking 30 minutes/day). Similar results can be observed for the case study represented in figure \ref{Figure_5c_MPC}.

\noindent \textbf{Case (1-d): Multiple Intervention - Varying Risk of Acquiring a New Condition:}
Figure \ref{Figure_5d_MPC} shows similar experimentation, but this time the proposed algorithm is deployed at two separate instances of time, i.e., at year 4 and year 8. The MPC determines that changing eating habits (start eating healthy) at year 4 and exercise habits (start exercise) while maintaining their healthy eating habits can reduce the probability of hyperlipidemia, high blood pressure, and obesity. After the second intervention in the following years, the likelihood of obesity remains constant compared to the original increasing probability. 

\noindent From this analysis, we can also determine that early intervention (figure \ref{Figure_5a_MPC}) can help patients better manage their health conditions. It is to mention that we assumed pessimistically that after each intervention, the patients stopped following the provided suggestions after a year.

In the second case scenario we assume a patient falls into the age group of $36 - 40$, doesn't follow a healthy eating habit, and doesn't perform any exercise.

\noindent \textbf{Case (2-a): Single Intervention - Low Risk of Acquiring a New Condition:} Figure \ref{Figure_6a_MPC} an intervention at year 2 is performed.
The MPC algorithm performs an online optimization and suggest changing the exercise regiment (starting exercise) and food habit (start eating healthy) can reduce the probability of hyperlipidemia and high blood pressure. 
The probability of other conditions remained unchanged due to this change. 

\noindent \textbf{Case (2-b \& c): Single Intervention - High Risk of Acquiring a New Condition:}
Figure \ref{Figure_6b_MPC} shows an intervention at year 3 is performed. The MPC algorithm again finds that changing the exercise regiment (starting exercise) and food habit (start eating healthy) can reduce the probability of hyperlipidemia, high blood pressure, and obesity. 
Similar results can be seen for the case study represented in figure \ref{Figure_6c_MPC}. 

\noindent \textbf{Case (2-d): Multiple Intervention - Varying Risk of Acquiring a New Condition:}
Figure \ref{Figure_6d_MPC} shows similar experimentation, but this time the proposed algorithm is deployed at two separate instances of time, i.e., at year 4 and year 8. While the algorithm identifies that changing the exercise and food habits at year 4 helped, the second intervention doesn't affect future probabilities.
After the second intervention in the following years, the probability of obesity rises at a low gradual rate compared to the original increasing likelihood. \\
It is to be mentioned, for both case studies, we pessimistically assumed that after each intervention, the patients stopped following the provided suggestions after a year.

\subsection{Limitation}
A potential shortcoming of the this study is that the study dataset doesn't allow testing of lifestyle changes in reality, which makes the validation of the proposed MPC algorithm difficult. However, we used the proposed model to compare the actual trajectories under no preventive measures (functional continuous time Bayesian network) with the predicted trajectories given the suggested behavioral changes to provide insight into the control and management of MCC. We also validated the suggestions made by the proposed approach by our medical team. 
%%%%%%%%%%%%%%%%%%%%%%%%%%%%%%%
Another limitation of the the study originates from the fact that, patients whose data was not maintained over the focused years were omitted. Aside from death, drop out can result from not requiring care or receiving care from other health providers, which typically happens to healthier patients and those who have health insurance. Consequently, the restricted population whose incomplete data are omitted, can be biased toward healthier patients and those with health insurance, which affect the predictability of the conditions.
%%%%%%%%%%%%%%%%%%%%%%%%%%%%%%%
The other limitation of this work is that the proposed behavioral changes suggested by the MPC algorithm are based on a convex objective function and may not consider an individual's complex/nonlinear scenarios. However, this problem is unlikely because convex optimization provides reasonably good results for short intervals, and the proposed MPC revisit/update its decisions for each unit time interval (recursive).

%\section{Bibliography styles}

%There are various bibliography styles available. You can select the
%style of your choice in the preamble of this document. These styles are
%Elsevier styles based on standard styles like Harvard and Vancouver.
%Please use Bib\TeX\ to generate your bibliography and include DOIs
%whenever available.

%Here are two sample references: 
%\cite{Fortunato2010}
%\cite{Fortunato2010,NewmanGirvan2004}
%\cite{Fortunato2010,Vehlowetal2013}

\section{Conclusion}
\label{Section: Conclusion}
This study proposes an integrated model predictive control functional continuous time Bayesian network to create an online estimation and decision making procedures for MCC networks. The proposed approach examine the impacts of modifiable risk-factors and identify the optimal change in the lifestyle behaviors along with their optimal timing on the emergence of MCC trajectory. 
The proposed model can further be personalized to make more/less aggressive decision by modifying the model constraints depending on patients risk factors and and pre-existing health conditions. We tested the proposed approach based on the live dataset of Cameron County Hispanic Cohort (CCHC) including 385 mainly Mexican Americans adults with five chronic conditions (DI, OB, HP, HL, CI), four socio-demographic factors (age, gender, education, marital status) and four lifestyle behavioral factors (diet, exercise, tobacco use, alcohol use).
We investigated multiple scenarios based on single and multiple interventions, as well as low and high risk of acquiring a new condition. 
The simulated trajectories generated by the proposed model reveals significant improvement in the risk of new MCC emergence under several tested scenarios, following the proposed behavioral changes.  In conclusion, the proposed model can support clinicians making proactive plans for patients to decrease the risk of developing new chronic conditions. Furthermore this model can be deployed in wearable devices to monitor and provide assistance to patients with high risk-factors which is planned for our future work.
\\

\section*{Acknowledgment}
The authors would like to thank the cohort team, particularly Rocío Uribe, who recruited and interviewed the participants. Marcela Morris, BS, and Hugo Soriano and their teams for laboratory and data support respectively; Norma Pérez-Olazarán, BBA, and Christina Villarreal, BA for administrative support; Valley Baptist Medical Center, Brownsville, Texas, for providing us space for our Center for Clinical and Translational Science Clinical Research Unit is located; and the community of Brownsville and the participants who so willingly participated in this study in their city. This study was funded in part by Center for Clinical and Translational Sciences, National Institutes of Health Clinical and Translational Award grant no. UL1 TR000371 from the National Center for Advancing Translational Sciences. This study was also partially funded by National Institutes of Health (NIH/NIGMS) Award grant no. 1SC2GM118266-01.

\bibliographystyle{model1-num-names}
%\bibliographystyle{cas-model2-names}

\bibliography{cas-refs}


% Author biography without author photo.
% Author biography. Author biography. Author biography.
% Author biography. Author biography. Author biography.
% Author biography. Author biography. Author biography.
% Author biography. Author biography. Author biography.
% Author biography. Author biography. Author biography.
% Author biography. Author biography. Author biography.
% Author biography. Author biography. Author biography.
% Author biography. Author biography. Author biography.
% Author biography. Author biography. Author biography.

% \bio{figs/pic1}
% Author biography with author photo.
% Author biography. Author biography. Author biography.
% Author biography. Author biography. Author biography.
% Author biography. Author biography. Author biography.
% Author biography. Author biography. Author biography.
% Author biography. Author biography. Author biography.
% Author biography. Author biography. Author biography.
% Author biography. Author biography. Author biography.
% Author biography. Author biography. Author biography.
% Author biography. Author biography. Author biography.

% \bio{figs/pic1}
% Author biography with author photo.
% Author biography. Author biography. Author biography.
% Author biography. Author biography. Author biography.
% Author biography. Author biography. Author biography.
% Author biography. Author biography. Author biography.

\end{document}